ANNEXURE – 1

# Medical Ministrations through Web Scraping

**Niketha Sabesan ,Nivethitha, J.N Shreyah ,**
**Pranauv A J , Shyam R ,**

***School of Computer Science and Technology (SCOPE)***
**VIT,Vellore,Tamil Nadu,India**

## I.INTRODUCTION

With the spread of the CoronaVirus pandemic and the dissemination of false information about the virus on multiple platforms, it is more important than ever to disseminate accurate information based on reliable medical research. Due to its user-friendly interface and accessibility, a chatbot can be useful in certain circumstances through web scraping. Users will be able to ask the chatbot any medical inquiry and through the dataset we will be web scraping data from the internet to know that they would get a response. The chatbot's ability to respond to questions will be further enhanced with the addition of generative question-answering, which might be applied to precision medicine. This approach to semantic document search will alter how a chatbot approaches and responds to a user's question by utilising a variety of retrieval algorithms. By building a pipeline of different IR approaches, including TF-IDF, BM-25, Dense Passage Retrieval, coupled with different language models. Our work addresses the issue of end to end open domain question answering systems.

## II.PROBLEM DEFINITION

As a real-world doctor would ask the patient more about any other symptoms they are having, by asking the co-occurring symptoms with the ones which were initially stated by the patient, we will apply this to the proposed system. The user is prompted till they have marked all the symptoms. By doing this we have incorporated a kind of approach to simulate the real-world interaction between patient and the medical practitioner.

These symptoms are then given as input to the trained model to make predictions. Top 10 most probable detected diseases are shown to the user with their respective independent probabilities or appropriate score. From the list of top diseases, the user can select the index of disease to know more details about any disease along with the treatment recommendation.

## III. LITERATURE REVIEW

*1)* Recurrent Neural Networks With TF-IDF Embedding Technique for Detection and Classification in Tweets of Dengue Disease by samina amin1 , m. irfan uddin 1 , saima hassan , atif khan , nidal nasser , abdullah al harb , and hashem alyami .The researchers explains the model that extracts the presence of patients infected with Dengue disease based on tweets only and decides whether it is a general discussion about the disease, and no one is actually infected, or people are actually infected with that disease. This paper uses different machine/deep learning approaches to utilise tweets data for automatic and efficient disease detection.

*2)* Detecting Liver Cancer from Online Search Logs by Tanvir Anjum Joarder,Boshir Ah,AHM Sarowar Sattarmed they proposed a method to organise online search log data to detect liver cancer early. They have used different machine learning techniques for this purpose, estimated the performance of SVM for different kernel scale values, and then compared the performance of all techniques on our dataset.

*3)* Risk Diagnosis and Mitigation System of COVID-19 Using Expert System and Web Scraping by Mohammad Robihul Mufid, Arif Basofi, Saniyatul Mawaddah the author explains the System to provide the latest information about the development of the COVID-19 case, Create a mobile application that can help people to diagnosis the early symptoms that people know risk indicator smitten by COVID- 19 with expert system rule-based method.

*4)* Healthcare Chatbot using Natural Language Processing by Papiya Mahajan, Rinku Wankhade, Anup Jawade, Pragati Dange, Aishwarya Bhoge .A Voice recognition chatbot has been developed that answers the user's query based on the questions asked by the user. The chatbot will take input from the user and process it with the algorithm and display the information as information about the disease. It uses the tf and idf algorithm to determine the score of relative importance of words and cosine similarity to measure the similarity of the words

*5)*Chatbot for Healthcare System Using Artificial Intelligence by Lekha Athota; Vinod Kumar Shukla; Nitin Pandey; Ajay Rana explains
the chatbot UI gets the user query and after that sends it to the chatbot application by In the chatbot application, the literary experiences pre-processing steps incorporate tokenization where the words are tokenized, feature extraction depends on n- gram, TF-IDF, and cosine likeness. The application uses n-gram for text compression using bigram and trigram for faster execution of the query and the Cosine similarity is used to check the similarity between two sentences.Using this the chatbot retrieves answers consequently for the questions.

*6)* Cosmetic Skin Type Classification Using CNN With Product Recommendation byArya kothari, Dipam shah, Taksh soni, Sudhir dhage . Most of the research papers make use of the colours and edges present in the images, and fall prey to any external noise, thereby giving false positives.

The literature available on cosmetic product recommendation does not provide a solution to suggest a skin care product based on an input of a facial image.

*7)* Better Understand Rare Disease Patients' Needs by Analysing Social Media Data byPapiya Mahajan1, Rinku Wankhade, Anup Jawade, Pragati Dange4, Aishwarya Bhoge have taken a Case Study of Cystic Fibrosis The aim is to analyse rare disease related posts from Reddit, one popular social media platform to reveal rare disease patients' needs based on hidden topics to be identified.

*8)* A Disease Identification Algorithm for Medical Crowdfunding Campaigns: Validation Study by Steven S Doerstling , Dennis Akrobetu: An international team of researchers has identified 11 disease categories in web-based medical crowdfunding campaigns with high precision and accuracy. Web scraping was used to collect data on medical crowdfunding campaigns.

*9)* Artificial intelligence for target symptoms of Thai herbal medicine by web scraping by Chairote Yaipraserta and Gorawit Yusakulb explains how a web scraping technique can be used to effectively get THM knowledge from huge data on the Internet.To pinpoint herbal properties and target symptoms, ML was utilised. Numerous symptoms may be combined to suggest therapeutic benefits or the opposite.

10) A Web scraped Skin Image Database of Monkeypox, Chickenpox, Smallpox, Cowpox, and Measles by Towhidul Islam, Mohammad Arafat Hussain, Forhad Uddin Hasan Chowdhury B. M. Riazul Islam explains the rarity of Monkeypox before the current outbreak further created a knowledge gap among healthcare professionals around the world. The lack of Monkeypox skin image data is making the bottleneck of using machine learning in Monkeypox detection from patient skin images.

## IV. PROPOSED SOLUTION

Diseases — Diseases are scraped from the National Health Portal of India, developed and maintained by the Centre for Health Informatics (CHI). This is combined with a predefined list of diseases to account for more diseases in the final prepared dataset.

Symptoms — Symptoms are scraped using a script that uses the Google Search package to perform searching and fetch the disease's Wikipedia page among the various search results obtained. The HTML code of the page is processed to fetch the symptoms of the disease using the 'info box' available on the Wikipedia page.

Pre-processing of Data set and Solution Sketch The scraped symptoms are pre-processed to remove similar symptoms with different names (For example, headache and pain in the forehead). To do so, symptoms are expanded by appending synonyms of terms in the symptom string and computing Jaccard Similarity Coefficient for each pair of symptoms.Symptom Expansion using Synonyms: Each symptom is expanded by appending a list of its synonyms. The synonyms are taken from thesauras.com (https://www.thesaurus.com/) and Princeton University's WordNet (https://wordnet.princeton.edu/) available in Python

Prediction using TF.IDF and Cosine Similarity Model: By calculating the TF and IDF of the dataset's symptoms, the TF.IDF score model is trained. The number of times a disease's symptom occurs is measured by TF (term frequency). The count of the symptom across all diseases is known as DF (document frequency)

## V. IMPLEMENTATION

The implementation is that symptoms are entered by the user in plain and everyday language in the chatbot. The user is then given the option to select their symptoms after query expansion utilising each symptom's synonyms and matching with symptoms contained in the dataset..Data exploitation is done by scrapping the dataset through: Diseases — The Centre for Health Informatics built and maintains the National Health Portal of India, from which diseases are scraped (CHI) and Symptoms — Using a script that searches Google and retrieves the disease's Wikipedia page, symptoms are extracted. The user receives prompts until they have noted every symptom. Then The user is presented with the top 10 most likely discovered diseases together with the corresponding independent probabilities or pertinent scores..Finally,The scraped symptoms are pre-processed to remove similar symptoms with different names. To do so, symptoms are expanded by appending synonyms of terms .

## VI.RESULT

This project is used to develop an open-source medical information retrieval chatbot that utilizes a transformer-based Reader and best practices in information retrieval to provide quick and effective answers from a large corpus of health-related material. The study compares cutting-edge dense passage retrieval techniques with traditional sparse retrieval techniques like Tf-If and Cosine Similarity .

## VII.SCREENSCHOT

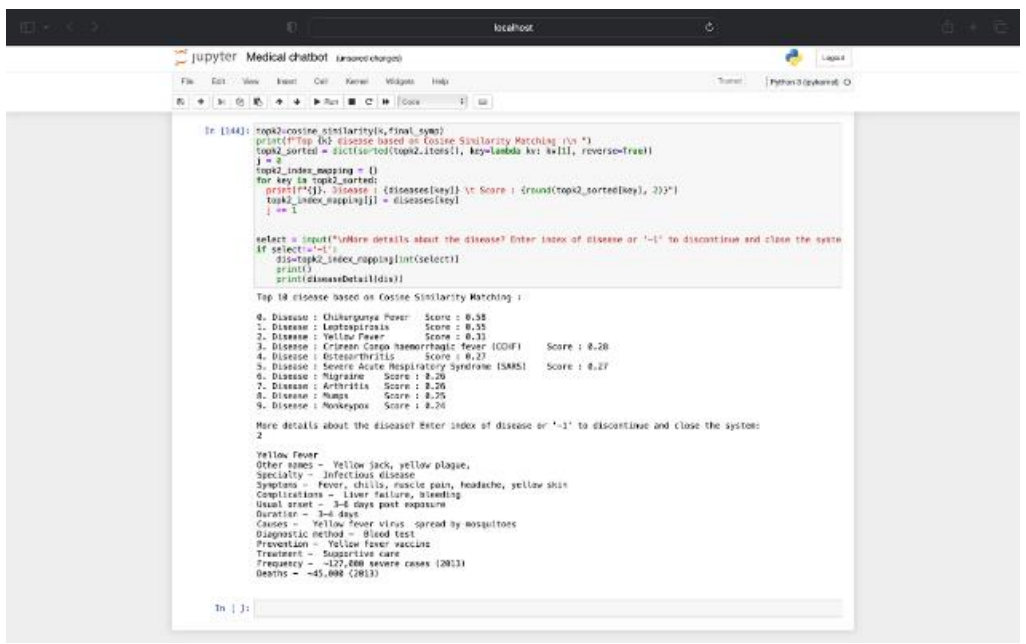

## VII.CONCLUSION

This project describes two techniques used in a medical information retrieval system. The first technique involves expanding symptoms by appending a list of their synonyms, which are obtained from thesaurus.com and Princeton University's WordNet using Python. The second technique involves using the TF.IDF score model to predict diseases based on symptoms. The TF.IDF score model is trained by calculating the term frequency (TF) and inverse document frequency (IDF) of symptoms in the dataset, and then measuring the number of times a disease's symptom occurs using TF and the count of the symptom across all diseases using DF.